\documentclass[twocolumn,trackchanges]{aastex7}
\usepackage{xcolor}
\usepackage{array, booktabs, adjustbox, siunitx}
\usepackage{amsmath}
\usepackage{cleveref}
\crefname{figure}{Figure}{Figures}
\crefname{section}{Section}{Sections}
\crefname{table}{Table}{Tables}
\crefname{algorithm}{Algorithm}{Algorithms}
\crefname{equation}{Equation}{Equations}
\crefname{section}{Section}{Sections}
\crefname{appendix}{Appendix}{Appendices}
\crefname{appendixsec}{Appendix}{Appendices}  
\crefformat{appendixsec}{#2Appendix~#1#3}     

\usepackage{textcomp}
\usepackage{enumitem}
\usepackage{subcaption}
\usepackage{booktabs}
\usepackage{graphicx}

\begin{document}

\title{Few-Shot Prediction for Pulsar Noise with Long Short-Term Memory Network}

\author[sname='Tang']{Qingye Tang}
\affiliation{Sichuan University, College of Computer Science}
\email{2022141520179@stu.scu.edu.cn}

\author[sname='An']{Dechao An}
\affiliation{Sichuan University, College of Physics}
\email{2023141220008@stu.scu.edu.cn}

\author[sname='Peng']{Haoran Peng}
\affiliation{Sichuan University, College of Computer Science}
\email{2022141520181@stu.scu.edu.cn}


\author[0000-0002-1611-1114, sname='Ouyang']{Yuqi Ouyang}
\altaffiliation{Corresponding author}
\affiliation{Sichuan University, College of Computer Science}
\email[show]{yuqi.ouyang@scu.edu.cn}


\begin{abstract}

This work proposes a novel solution to predict pulsar timing residuals with limited data, addressing
the critical challenge of data scarcity across spin-frequency subgroups of millisecond pulsars in PTA datasets. The proposed solution applies a Long Short-Term Memory (LSTM) network optimized using the model-agnostic meta-learning algorithm, enabling rapid adaptation to new frequency domain by fine-tuning the LSTM network with only a few-shot of ground truth timing residuals. Particle swarm optimization algorithm is also used for automatic hyperparameter optimization, leading to improved prediction accuracy. Our solution, evaluated on the second data release of the International Pulsar Timing Array (IPTA), demonstrates robust generalization with accurate predictions in three metrics across high-frequency test frequency domains, while requiring only 10\% of the timing residuals from these domains for model fine-tuning. Furthermore, our lightweight structure only costs 16.86 MB CPU memory and 18 milliseconds for single-step residual prediction. All these characteristics make our solution highly suitable for real-world applications, where effective and real-time predictions of pulsar timing residuals are essential—particularly in resource-constrained environments  with limited computational power, memory, or energy availability.

\end{abstract}



\section{Introduction}
Pulsar timing is a critical technique for detecting gravitational waves in the nanohertz–microhertz frequency regime by characterizing timing residuals derived from time-of-arrival measurements (TOAs) \citep{RN48,RN49,RN50}. It relies on accurate TOAs of pulses from millisecond pulsars, which are recorded with uncertainties arising from instrumental effects and radio frequency interference (RFI) \cite{RN52}. Timing residuals, defined as the differences between observed and predicted TOAs,  arise due to deviations of the timing model parameters from their true values as well as various noise processes that must be characterized to  ensure robust and statistically consistent inference of timing model parameters. These residuals reflect the combined influence of imperfect timing model parameters and stochastic noise processes that perturb pulse arrival times and therefore form the basis for assessing the fidelity of the timing model. A key difficulty is that these contributions can mask or become covariant with astrophysical signals embedded in the TOAs. For instance, perturbations induced by nanohertz gravitational waves from the inspiral and merger of supermassive black holes can be covariant with rotational irregularities, interstellar medium variations, or residual instrumental noise \cite{R201}. Consequently, the accurate prediction of timing residuals establishes a dynamic empirical baseline for pulsar observations. This serves as a vital practical utility for the real-time identification of anomalies, such as unexcised radio frequency interference (RFI), instrumental systematics, or sudden interstellar medium (ISM) events \cite{RN52}, prior to the execution of computationally intensive offline analyses such as Gaussian Processes \cite{RN41, RN43}. Also, these residuals exhibit structure on timescales determined by stochastic noise processes and long-term timing irregularities. Each pulsar emits pulses at a rate given by its rotation frequency, typically ranging $\sim$100~Hz to over 600~Hz for millisecond pulsars and varying across the pulsar timing array population. To enable few-shot domain adaptation within the MAML framework, we partition pulsars into frequency domains based on their spin frequency as a practical grouping strategy. This allows the model meta-trained on well-sampled lower-frequency domains to rapidly adapt to higher-frequency domains with limited observations, thereby improving timing model fidelity and supporting the detection of astrophysical signals such as gravitational waves.


\begin{figure*}[!t]
    \centering
    \includegraphics[width=\linewidth]{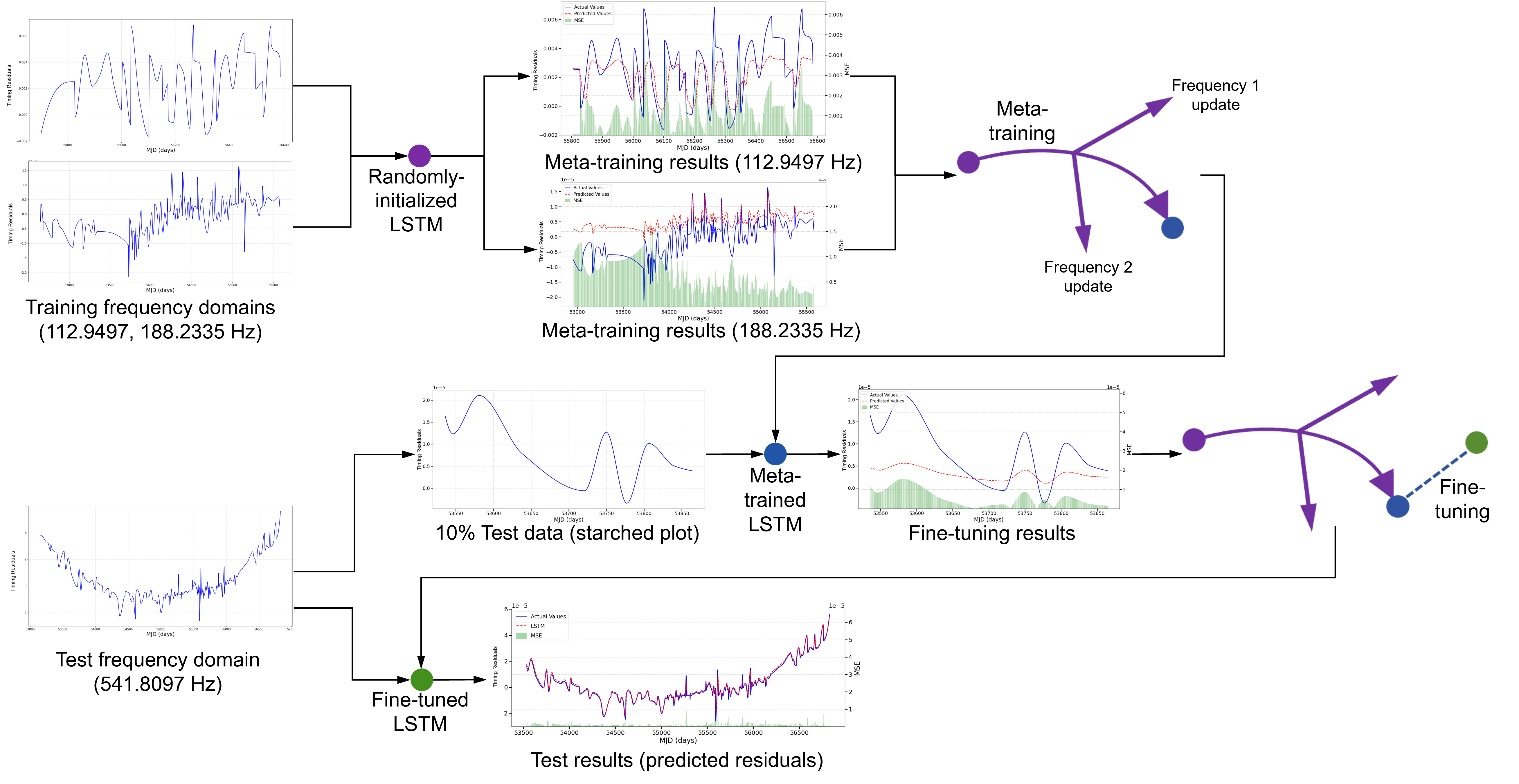}
    \caption{An overview of our solution including: meta-training, fine-tuning and testing. The example of test ratio between model fine-tuning and performance evaluation is $1:9$, this ratio can be further adjusted.}
    \label{fig:FewShot}
\end{figure*}

The task of noise modeling has been a challenge in the pulsar timing community, with efforts evolving over decades to refine techniques and address persistent flaws. Initially, \cite{RN24} pioneered basic statistical approaches to characterize timing residuals, laying a foundation for noise subtraction; however, their method had difficulty capturing the the power-law spectrum of red noise, which exhibits stronger power at low Fourier frequencies. Building on this, \cite{RN03} introduced more robust data preparation pipelines, establishing pulsar timescales through high-precision observations, while was still challenged by the complexity of low-frequency noise components. \cite{RN25} identified pervasive low-frequency timing irregularities, exposing the limitations of prior frameworks and prompting a shift toward more advanced methodologies. Recently, \cite{RN04} developed a frequency analysis framework within dedicated software, enabling simultaneous fitting of noise and timing parameters to improve accuracy. In addition to these contributions, Gaussian processes have emerged as a cornerstone of pulsar noise analysis, offering a flexible, non-parametric approach to model both white and red noise components, significantly improving the precision of noise parameter estimation \cite{RN40,RN41,RN42,RN43}. Furthermore, Kalman filters have recently been adopted as a method for real-time, adaptive noise modeling, particularly effective for handling non-stationary noise processes \cite{RN44}.

While representing a significant advancement, previous approaches remain limited, due to the high $O(n^3)$ computational complexity of standard Gaussian Processes and the rigid parametric assumptions often required to make noise modeling computationally tractable for large datasets \citep{RN41}, this underscores the need for innovative approaches. In the new era, deep learning methods have shown excellent capability in data modeling and analysis \cite{RN14,RN35,RN36,RN37}. In particular, the Long Short-Term Memory networks (LSTMs), introduced by \cite{RN14}, excel in predictive modeling tasks by capturing long-term dependencies within sequential data \cite{RN39}. This makes LSTMs ideal for pulsar timing, where temporal correlations arising from all stochastic contributions to the timing residuals, including achromatic red noise from spin irregularities, chromatic ISM effects, and instrumental noise, allow for sequential prediction, aligning closely with our task of modeling the total residuals. However, despite the suitability of LSTM networks for the prediction task, two challenges still remain: data scarcity and hyperparameter optimization of the deep learning models. While radio telescopes provide pulsar observations, data scarcity exists for newly discovered or high-frequency MSPs due to limited telescope time and instrumental constraints \cite{RN07,RN08,RN09}. Deep learning models, which are based on large labeled datasets, typically suffer from data scarcity \cite{RN10}. To address this, a few-shot learning strategy is proposed. In this strategy, pulsars are first grouped into domains based on spinning frequencies. A prediction model is then trained on well-represented domains and adapted to data-scarce ones. This can be achieved using model-agnostic meta-learning (MAML), an optimization algorithm compatible with any gradient-based deep learning model, enabling rapid adaptation with minimal data \cite{RN38}. As for the challenge of hyperparameter optimization, traditional grid search in deep learning models is often costly, time-consuming, and requires human supervision. To improve efficiency, the Particle Swarm Optimization (PSO) algorithm \cite{RN06} is incorporated to automate hyperparameter tuning, enhancing prediction accuracy and model robustness. 

Prior to this work, the standard noise modeling method using Gaussian processes \cite{R202}, here we summarize the qualitative comparison between the two approaches: The Gaussian-process (GP) framework commonly used in pulsar-timing noise analysis models timing residuals as sums of stochastic processes whose covariance functions encode physically motivated priors (e.g., power-law red noise, frequency-dependent DM, and inter-pulsar Hellings–Downs correlations). This yields principled marginalization, component separation and predictive uncertainty quantification that are particularly useful for detection or attribution tasks. However, its original form incurs $O(n^{3})$ scaling in the number of TOAs, practical PTA analyses therefore rely on low-rank or Fourier expansions and specialized samplers to make full-noise Bayesian inference tractable. GP methods are for the situation when principled uncertainties and physical component separation are required (for example, in gravitational-wave searches searches). By contrast, our approach is fully data-driven and predicatively few-shot: we meta-train a bidirectional LSTM on uniformly resampled (1-day grid) residual sequences across well-sampled frequency domains, and fine-tune with only a small fraction (10\%) of data from a target domain. Additionally, PSO is incorporated for automatic hyperparameter selection. These enable our model to capture potentially nonlinear and nonstationary structures without prescribing a covariance family, while producing fast, lightweight single-step predictions once meta-trained (16.9 MB and 18 ms per step in our tests). Compared with GP framework, our solution excels when fast and accurate few-shot predictive performance and low computational footprint are the priority. To sum up:

This work proposes a novel solution to address the data scarcity challenge in pulsar noise analysis. First, MAML is applied for few-shot learning, where the model is first trained with multiple pulsars across low-frequency domains categorized by spinning frequencies, and then fine-tuned with a limited amount of data from a targeted high-frequency domain, achieving domain adaptation with accurate prediction results. Furthermore, the PSO algorithm is utilized to provide an automated solution for hyperparameter optimization, further enhancing the prediction performance. The main contributions of this work are summarized as follows: 
\begin{figure*}[!t]
    \centering
    \includegraphics[width=\linewidth]{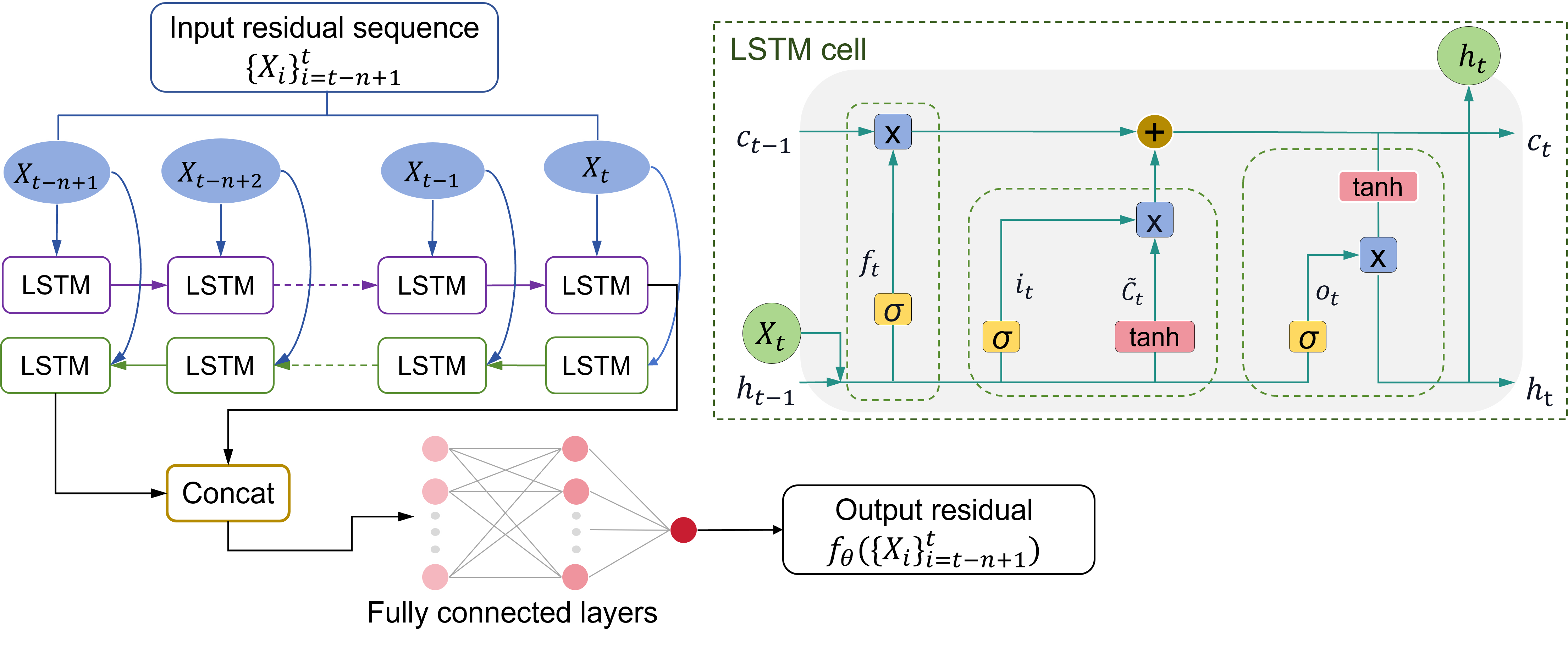}
    \caption{The computational process of our prediction task, where $\{{X}_{i}\}_{i=t-n+1}^{t}$ is the input residual} sequence; $f_{{\theta}}\left(\{{X}_{i}\}_{i=t-n+1}^{t}\right)$ is the next-step output of timing residual with $\theta$ denoting network parameters; $X_{t}$ is cell input; $F_{t}$, $i_{t}$, $o_{t}$ denote the output of the forget gate, input gate, output gate, respectively; $c_{t-1}$, $c_{t}$ are the cell states; $h_{t-1}$, $h_{t}$ are the hidden states.
    \label{fig:LSTM}
\end{figure*}
\begin{enumerate}[label=\arabic*)]
    \item Our work addresses the data scarcity issue in pulsar noise analysis, achieving accurate predictions across multiple frequency domains in the second data release of the International Pulsar Timing Array (IPTA) \cite{RN46}, while requiring only a few data from the targeted frequency domains.
    \item A fully automated solution is provided with automatic hyperparameter optimization, leading to further enhanced prediction accuracy and model robustness.
    \item Real-time efficiency is achieved with a lightweight model, requiring only 16.9 MB of CPU memory for an inference speed of 18 milliseconds. Our prediction accuracy, efficiency, and generalizability make our solution well suited for real-world applications. Comprehensive experimental analysis is presented to validate our core functionalities.
\end{enumerate}

The remainder of this paper is organized as follows. In Section \ref{sec:Solution}, we present our proposed solution, detailing the bidirectional LSTM architecture, the application of the MAML framework for few‐shot domain adaptation, and the integration of the PSO algorithm for automatic hyperparameter tuning. Section \ref{sec:Experiment} describes the experimental setup, including the IPTA DR2 dataset, implementation details, evaluation metrics, and comprehensive performance results  across multiple test domains.

\section{Methods} 
\label{sec:Solution}
In this work, unless otherwise specified, references to frequency domains denote groups of pulsars categorized
by their rotational frequencies as a practical partitioning strategy for MAML domain adaptation on the IPTA DR2
MSP ensemble. This is distinct from radio observing frequencies (MHz-GHz) and Fourier frequencies (nHz).

Depicted in \cref{fig:FewShot}, our solution includes three processes: meta-training, fine-tuning, and testing. In meta-training, timing residuals from low-frequency domains (112.9497 Hz and 188.2335 Hz) are used to form cross-frequency prediction tasks, leading to a coarse LSTM initialization named meta-trained LSTM, which is further updated in the fine-tuning stage with only a few timing residuals from an unseen high frequency domain (10\%, 541.8097 Hz in our example), performing fast and accurate domain adaptation in a few-shot setting.

LSTM selectively retains or discards information through a gate system in its cell structure, effectively addressing the gradient vanishing and exploding issues often encountered when processing long sequences \cite{RN15}. Bidirectional LSTMs capture temporal dependencies in both forward and backward directions, allowing for improved integration of contextual information and more accurate modeling of temporal patterns in sequential data \cite{RN33,RN34}. Based on this, our prediction task, as depicted in \cref{fig:LSTM}, starts from a bidirectional LSTM processing the historical timing residual sequence $\{X_i\}_{i=t-n+1}^{t}$ which represents resampled timing residuals. Since original pulsar observations are recorded at irregular intervals, we apply linear interpolation to map the observed residuals onto a uniform grid with a constant 1-day interval in Modified Julian Date (MJD). This sequence, processed with a window size of $n$ steps, is followed by concatenating the bidirectional hidden states, and mapping the concatenation through fully-connected layers to predict the next-step timing residual $f_{\theta}(\{X_i\}_{i=t-n+1}^{t})$, with $f_{\theta}(\cdot)$ denoting the forward computing process using trainable parameter set $\theta$. The timestep interval is one day of MJD.

\subsection{The LSTM Cell}
In \cref{fig:LSTM}, the LSTM layer contains multiple cells with common neural structure depicted for pattern analysis, highlighting the gate system that controls the flow of information. Such a gate system consists of three main components: the forget gate, the input gate and the output gate, explained next.

The forget gate decides information discarding from the cell state, presented as:
\begin{equation}
\centering
    F_t = \sigma \left( W_F\cdot\left[h_{t-1}, X_t\right] + b_F \right),
\end{equation}
where $W_F$ is the weight matrix, $b_F$ is the bias, $X_t \in \mathbb{R}^n$ is the input vector at time step $t$, $h_{t-1} \in \mathbb{R}^m$ is the previous hidden state, and $\sigma(\cdot)$ denotes the Sigmoid function. Thus, $F_t$ is a gating vector with real numbers bounded in $(0,1)$, determining the proportion of information to forget at each unit.

The input gate determines information addition to the cell state, as follows:
\begin{equation}
\centering
    i_t = \sigma \left( W_{i}\cdot\left[h_{t-1}, X_t\right] + b_i \right),
\end{equation}

\begin{equation}
\centering
    \tilde{C}_t = \tanh \left( W_{C}\cdot\left[h_{t-1}, X_t\right] + b_C \right),
\end{equation}
where $W_{i}$ and $W_{C}$ are the weight matrices; $b_i$ and $b_{C}$ are the bias terms; $\tanh(\cdot)$ denotes the hyperbolic tangent function. Therefore, $i_t$ is the input gate vector with real numbers bounded in $(0,1)$, and $\tilde{C}_t$ is the candidate cell state vector with values bounded in $(-1,1)$.

The cell state is computed as a weighted addition between the previous cell state and the candidate cell state:
\begin{equation}
\centering
    c_t = F_t\cdot c_{t-1} + i_t \cdot\tilde{C}_t,
\end{equation}
where $c_{t-1}$ is the previous cell state. $c_t$ integrates both the retained and updated information.

In the end, the output gate computes the hidden state $h_t$ based on the cell state $c_t$, as follows:
\begin{equation}
\centering
    o_t = \sigma \left( W_{o}\cdot\left[h_{t-1},X_t\right] + b_o \right),
\end{equation}

\begin{equation}
\centering
    h_t = o_t \cdot \tanh \left( c_t \right),
\end{equation}
where $W_{o}$ is the weight matrix, $b_o$ is the bias, the hidden state $h_t$ is with real numbers bounded in $(-1,1)$. The design of gate system enables LSTM to extract and maintain long-term dependencies from time-series data.

\subsection{Model-Agnostic Meta-Learning}
\begin{figure}[!t]
    \centering
    \includegraphics[width=0.92\linewidth]{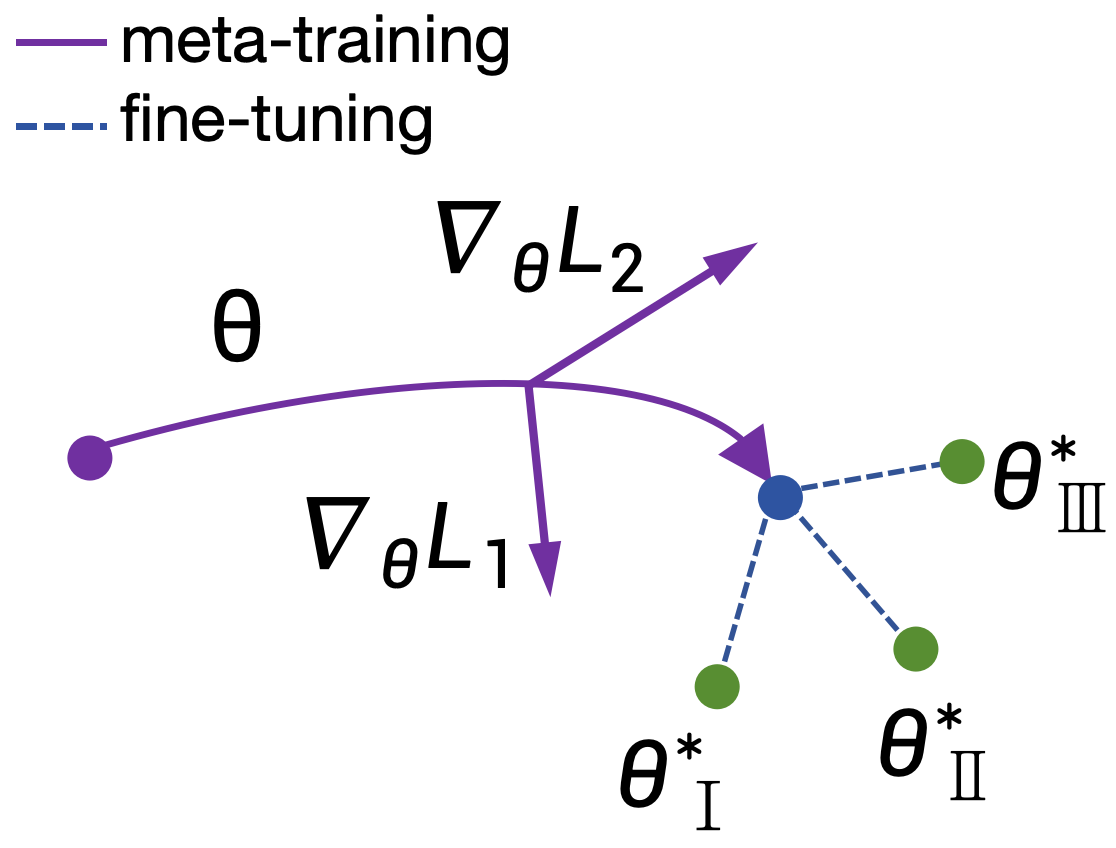}
    \caption{An example demonstration of the MAML learning strategy in our solution. $\nabla_{\theta} L_1$ and $\nabla_{\theta} L_2$ are the gradients computed from the prediction tasks respectively on 2 different frequency domains. Blue point denotes the LSTM initialization by meta-training. Each green point depict an optimal LSTM fine-tuned on each new frequency domain.}
    \label{fig:MAML-few}
\end{figure}

\cref{fig:MAML-few} depicts a demonstration of our MAML learning strategy. Based on parameter optimizations on multiple frequency domains, the meta-training process searches for a model initialization as a prior knowledge. Hence, instead of training from the ground for new frequencies, fast adaptation is enabled to each new frequency domain in fine-tuning process, such few-shot strategy addresses data scarcity issue through domain adaptation. All optimizations in the MAML strategy are based on gradient descent guided by mean squared error (MSE) loss computed from the residual prediction tasks, denoted as follows:

\begin{equation}
\centering
    L_{j}\left(f_{\theta}\right) = \frac{1}{N} \sum_{\{X_{i}\},y_{i}\sim \mathcal{D}_{j}} \bigg\Vert f_{\theta}\left(\{X_{i}\}\right)-y_{t+1} \bigg\Vert_{2}^{2},
\label{eq:MSE}
\end{equation}
where $\{{X}_{i}\},y_{t+1}\sim \mathcal{D}_{j}$ denotes that the the input $\{{X}_{i}\}$ and the ground truth timing residuals $y_{t+1}$ are sampled from the dataset $\mathcal{D}_{j}$ of the $j_{\mathrm
{th}}$ frequency domain; $\Vert\cdot\Vert_{2}$ computes the $l_2$ norm; $N$ denotes to the sample size.

The meta-training stage consists of residual prediction tasks from multiple frequency domains, adopting a nested optimization to learn an LSTM initialization, mathematically formulated as follows:

\begin{equation}
\centering
    \theta_{i+1}=\theta_{i}-\alpha\sum_{j}\nabla_{\theta}L_{j}\left(f_{\theta_{i}}\right), \,\,\, i\in[1,k-1]
\label{eq:MetaTrainInner}
\end{equation}

\begin{equation}
\centering
    \theta_{1}\leftarrow\theta_{1}-\beta\sum_{j}\nabla_{\theta_{1}}L_{j}\left(f_{\theta_{k}}\right),
\label{eq:MetaTrainOuter}
\end{equation}
where $\alpha$ and $\beta$ are learning rates for the inner optimization and outer optimization, respectively; the subscript $i$ denotes the step number of inner gradient descent, where $\theta_{1}$ indicates the LSTM initialization. \cref{eq:MetaTrainOuter} shows that optimizing the LSTM initialization $\theta_{1}$ in the outer loop depends on the resulted $\theta_{k}$ from the inner loop. Hence, gradient descent of the outer optimization requires computing the Hessian at the inner optimization, triggering the improved gradient and therefore facilitates fast model adaptation on new data domains. Note that to implement the nested optimization, data samples from each training frequency domain $j$ are partitioned into two sets for the inner and outer parts, respectively.

After meta-training, the fine-tuning process rapidly adapts the learned initialization to new frequency domains, as depicted by the blue dashed lines in \cref{fig:MAML-few}. Such adaptation runs with a small number of gradient steps, calculated from the prediction tasks associated with limited data from each test frequency domain.

\subsection{Particle Swarm Optimization}
In our solution, the PSO algorithm finds optimal hyperparameters that leads to the minimal loss value. To achieve this, the algorithm regards these hyperparameters as particle locations, then initiates and iteratively updates multiple particles' velocity and location, respectively as follows:

\begin{equation}
    v_{i} = \omega v_{i-1} + c_1 r_1\cdot(m_{\ast} - m_{i}) + c_2 r_2\cdot(M_{\ast} - m_{i}),
\label{eq:PSOv}
\end{equation}

\begin{equation}
    m_{i+1} = m_{i} + v_{i},
\end{equation}
where $v$ denotes particle velocity; $m$ means particle location; subscript $i$ is for the PSO iteration number; subscript $\ast$ defines current optimal locations with minimal loss, while $m_{\ast}$ and $M_{\ast}$ respectively denote particle's optimal location and the whole swarm's global optimal location, dynamically renewed with the ones associated with lower loss values; $\omega$ is the inertia weight; $c_{1}$ and $c_{2}$ are the learning factors; $r_{1}$, $r_{2}$ are random numbers that ensures searching diversity.

In \cref{eq:PSOv}, the velocity, i.e., the moving direction of each particle is computed based on the personal optima $m_{\ast}$ and the shared information of the global optima $M_{\ast}$, reflecting the natural swarm behavior of information sharing while exploring for resources. By covering a wide search space, such swam optimization strategy helps escape local minima, leading to more robust result \cite{RN16}. Following the iterative process, the PSO algorithm provides an automatic solution for hyperparameter setting. In this work, the PSO algorithm is applied to search for three key hyperparameters for the meta-training process: learning rate and gradient descent iteration number in the inner-optimization, and the learning rate in the outer-optimization.

\subsection{Comparison with Standard Noise Modeling Methods}
The Gaussian-process (GP) framework van Haasteren \& Vallisneri (2014) commonly used in pulsar-timing noise analysis models timing residuals as sums of stochastic processes whose covariance functions encode physically motivated priors (e.g., power-law red noise, frequency-dependent DM, and inter-pulsar Hellings–Downs correlations). This yields principled marginalization, component separation and predictive uncertainty quantification that are particularly useful for detection or attribution tasks. However, its original form incurs $O(n^{3})$ scaling in the number of TOAs, practical PTA analyses therefore rely on low-rank or Fourier expansions and specialized samplers to make full-noise Bayesian inference tractable. GP methods are for the situation when principled uncertainties and physical component separation are required (for example, in gravitational-wave searches searches).

By contrast, our approach is fully data-driven and predicatively few-shot: we meta-train a bidirectional LSTM on uniformly resampled (1-day grid) residual sequences across well-sampled frequency domains, and fine-tune with only a small fraction (10\%) of data from a target domain. Additionally, particle swarm optimization (PSO) is incorporated for automatic hyperparameter selection. These enable our model to capture potentially nonlinear and nonstationary structures without prescribing a covariance family, while producing fast, lightweight single-step predictions once meta-trained (16.9 MB and 18 ms per step in our tests). Compared with GP framework, our solution excels when fast and accurate few-shot predictive performance and low computational footprint are the priority.

\section{Experiment}
\label{sec:Experiment}
\subsection{Dataset}
In this work,  version B of IPTA DR2 dataset \cite{RN27} is used. In the dataset, long-term, high-precision pulsar timing observations from three constituent PTAs are integrated to enhance temporal diversity and global spatial coverage for improved sensitivity to nanohertz gravitational wave signals. The three PTAs are: the European Pulsar Timing Array (EPTA) \cite{RN17}, the North American Nanohertz Observatory for Gravitational Waves (NANOGrav) \cite{RN18}, and the Parkes Pulsar Timing Array (PPTA) \cite{RN19}. More specifically, the IPTA DR2 dataset includes timing observations from 49 millisecond pulsars (MSPs), which exhibit red noise characteristics. The data span a temporal baseline of up to 24 years, with MJD values ranging from 46436 (1986 January 06) to 56598 (2013 November 02), based on the earliest observations recorded in \cite{RN28} and the latest in \cite{RN29}. The selected MSPs, characterized by rapid rotation, have spin frequencies ranging from 62.3 Hz to 641.8 Hz \cite{RN27}. Low-frequency noise is dominated by long-term trends and interstellar medium effects, resulting in higher amplitude but smoother variations \cite{RN30}. Comparatively, high-frequency noise is more challenging to model due to the variance caused by instrumental and short-term astrophysical effects \cite{RN31,RN32}, forming the primary focus of domain adaptation for modeling timing residuals in our work.

\subsection{Implementation Details}
The timing residuals, as the difference between the actual TOA and the predicted TOA based on the timing model \cite{RN20,RN21}, are linearly rescaled to the $[0,1]$ range in this work. The bidirectional LSTM in our solution are designed with two stacked layers and a hidden size 200. To mitigate overfitting, dropout with rate 0.2 is applied on LSTM and LSTM output. In the end, fully connected layers are used to produce the output, the layer-wise neurons numbers are 100 and 1, with the first layer activated with ReLU function.

Pulsar timing residual sequences in the PTA sample exhibit variations across the observed range of spin frequencies,
which we use to define domains for cross-pulsar adaptation \cite{RN22,RN23}. To demonstrate our adaptation ability, we design the training on two low-frequency domains of 112.950 Hz and 188.234 Hz, while the fine-tuning and evaluations are set on three high-frequency domains of 607.670 Hz, 541.810 Hz, and 420.189 Hz. During meta-training, the LSTM network is optimized for 100 epochs, with timing residuals from each frequency domain divided into $10\%$ for inner optimization with learning rate $0.005$, and $90\%$ for outer optimization with learning rate $0.0005$. The PSO algorithm is applied to set learning rate and gradient descent iteration number in the inner-optimization, and the learning rate in the outer-optimization as $0.000927$, $7$ and $0.000914$, 
ranging between $10^{-4}$ and $10^{-2}$, $1$ and $10$, as well as $10^{-5}$ and $10^{-3}$,  respectively. 5 particles are initialized with a maximum iteration 10. The inertia weight of the PSO optimizer is set to 0.9. The two learning factors for social and individual components are configured as 0.5 and 0.3 respectively. For each test frequency domain, the first $10\%$ timing residuals is used for fine-tuning, while the remaining $90\%$ is for performance evaluation. Adam \cite{RN47} optimizers with the default setting are applied for all learning tasks.

\begin{figure*}[!t]
    \centering
    \includegraphics[width=1\linewidth]{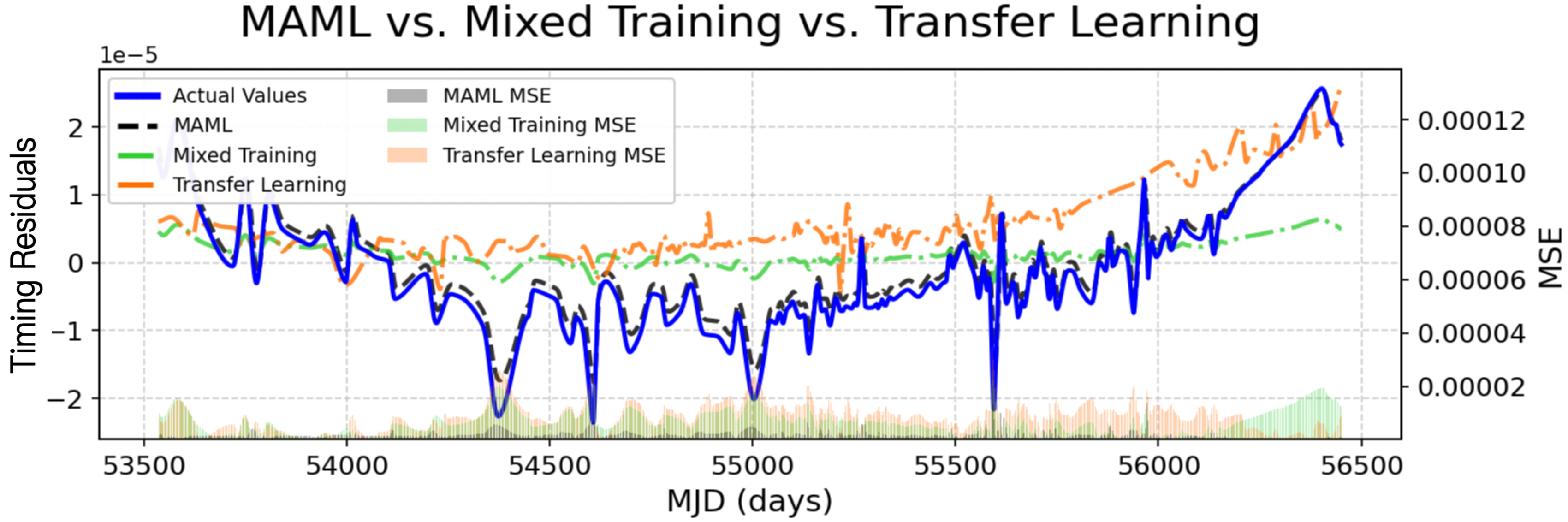}
    \caption{The predictions of timing residuals at test frequency 541.810, computed from mixed training (green curve), transfer learning (orange curve), or MAML (black curve) strategies.}
    \label{fig:MAML_comparative}
\end{figure*}

\subsection{Evaluation Metrics}
Various metrics are applied to evaluate and compare model performance, they are: mean absolute error (MAE), mean squared error (MSE), mean absolute percentage error (MAPE), and coefficient of determination ($R^2$), calculations are respectively defined as follows:

\begin{equation}
    \text{MAE} = \frac{1}{N} \sum_{\{X_{i}\},y_{i}\sim \mathcal{D}_{j}} \bigg| f_{\theta}\left(\{X_{i}\}\right)-y_{t+1} \bigg|,
\end{equation}

\begin{equation}
    \text{MSE} = \frac{1}{N} \sum_{\{X_{i}\},y_{i}\sim \mathcal{D}_{j}} \bigg\Vert f_{\theta}\left(\{X_{i}\}\right)-y_{t+1} \bigg\Vert_{2}^{2},
\end{equation}

\begin{equation}
    \text{MAPE} = \frac{1}{N} \sum_{\{X_{i}\},y_{i}\sim \mathcal{D}_{j}} \left| \frac{f_{\theta}\left(\{X_{i}\}\right) - y_{t+1}}{y_{t+1}} \right| \times 100\%,
\end{equation}

\begin{equation}
R^2 = 1 - \frac{\sum_{\{X_{i}\}, y_{i} \sim \mathcal{D}_{j}} \left( f_{\theta}\left(\{X_{i}\}\right) - y_{t+1} \right)^2}{\sum_{\{X_{i}\}, y_{i} \sim \mathcal{D}_{j}} \left( y_{t+1} - \bar{y} \right)^2},
\end{equation}

where $\{{X}_{i}\},y_{i}\sim \mathcal{D}_{j}$ denotes that the residual sequence $\{{X}_{i}\}$ and the ground truth $y_{t+1}$ is sampled from the $j_{\mathrm
{th}}$ frequency domain. More accurate predictions are those with lower MAE, lower MAPE, or higher $R^{2}$ values.

\subsection{Overall Performance}
The prediction results on three test frequencies are tabulated in \cref{tab:MAML_overview}, overall depicting high $R^2$ values. Especially the 0.9933 $R^2$ value at test frequency 541.810 Hz. Note that our solution is only fine-tuned with $10\%$ of the data from these large frequency domains. These findings indicate that the our solution can address the data scarcity issue with accurate prediction results. However, the functionalities of model components require further studies to validate, presented in the following sections.

\setlength{\tabcolsep}{10pt}
\begin{table}[!t]
    \centering
    \caption{The prediction results of timing residuals at three frequencies.}
    \label{tab:MAML_overview}
    \begin{tabular}{lcccccc}
        \toprule
        Frequency (Hz) & \multicolumn{3}{c}{metrics} \\
        \cmidrule(lr){2-4}
        &  MAE(\%) &  MAPE &  $R^2$ \\
        \midrule
        420.189 &$0.0000005$
 & 0.2736 & 0.9828 \\
        541.810 & 0.0001 & 0.2021 & 0.9933 \\
        607.678 & 0.0002 & 0.4210 & 0.9520 \\
        \bottomrule
    \end{tabular}
\end{table}

\subsection{Comparative Study of MAML}
To study the effect of our few-shot strategy, we compare MAML with mixed training and transfer learning strategies. In mixed training, the model is trained by combining the low-frequency training data with the 10\% high-frequency test data, while in transfer learning, the model is first trained with the training set and subsequently fine-tuned using the same 10\% test data. The prediction results are summarized in \cref{tab:MAML_comparative}, demonstrating that MAML achieves more accurate prediction results across all three test frequency domains, exhibiting lower MAE, lower MAPE, and higher $R^2$ values. Notably, at 420.189 Hz, MAML reduces the MAP from 0.0006\% from transfer learning to 0.0001\%, while significantly improving $R^2$ from -0.4155 to 0.9019. At 541.810 Hz, MAML maintains the lowest MAPE of 0.8069, substaintially reducing those achieved by mixed Training and Transfer Learning. At 607.678 Hz, MAML achieves an $R^2$ value of 0.8753, outperforming the 0.6708 attained by mixed training and 0.3545 achieved via transfer learning. These results conclusively show that conventional training strategies suffer from data scarcity, while the proposed few-shot MAML strategy addresses this issue with improved model generalization.

\begin{figure*}[!t]
    \centering
    \includegraphics[width=0.49\textwidth]{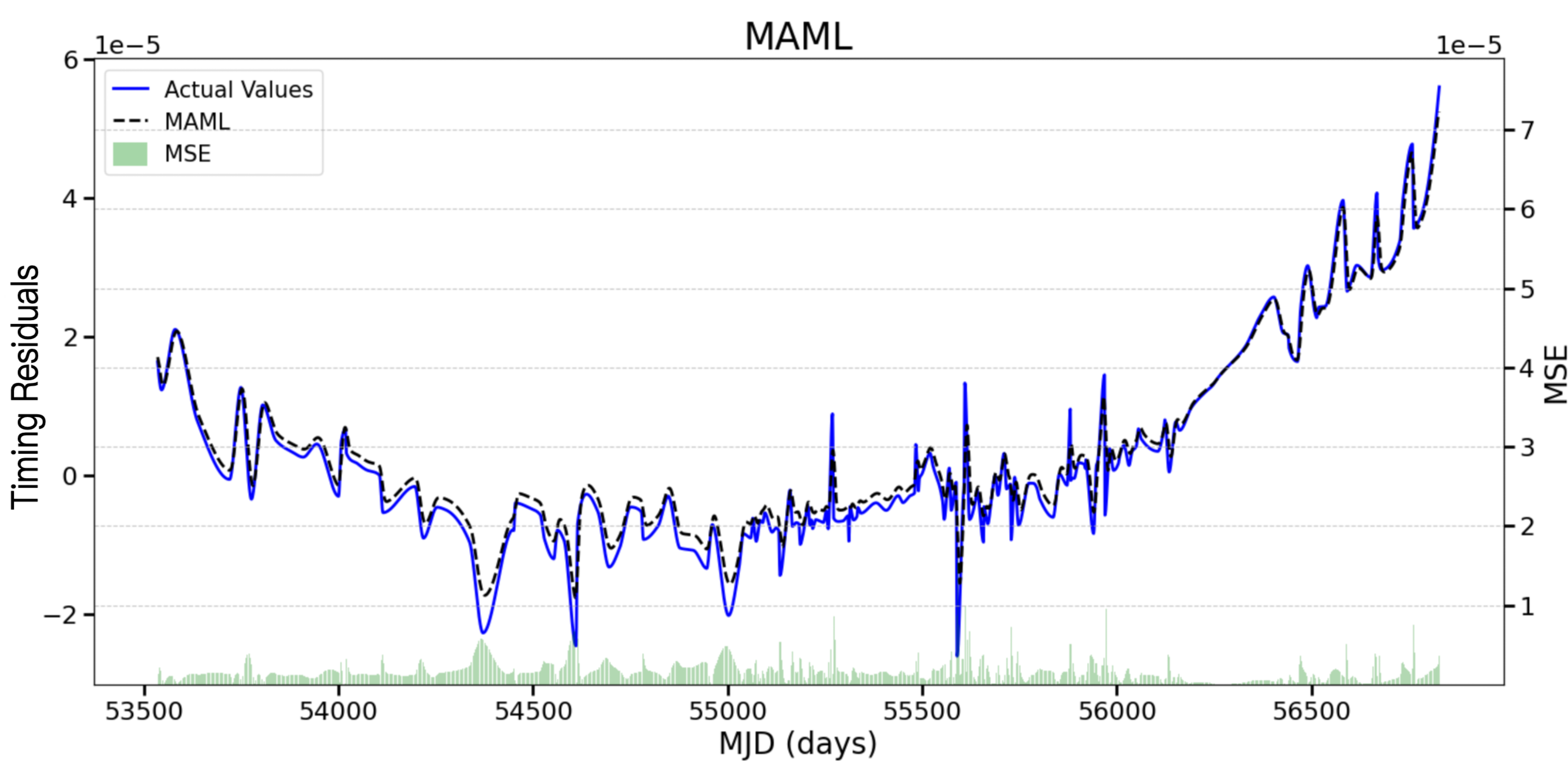} 
    \includegraphics[width=0.49\textwidth]{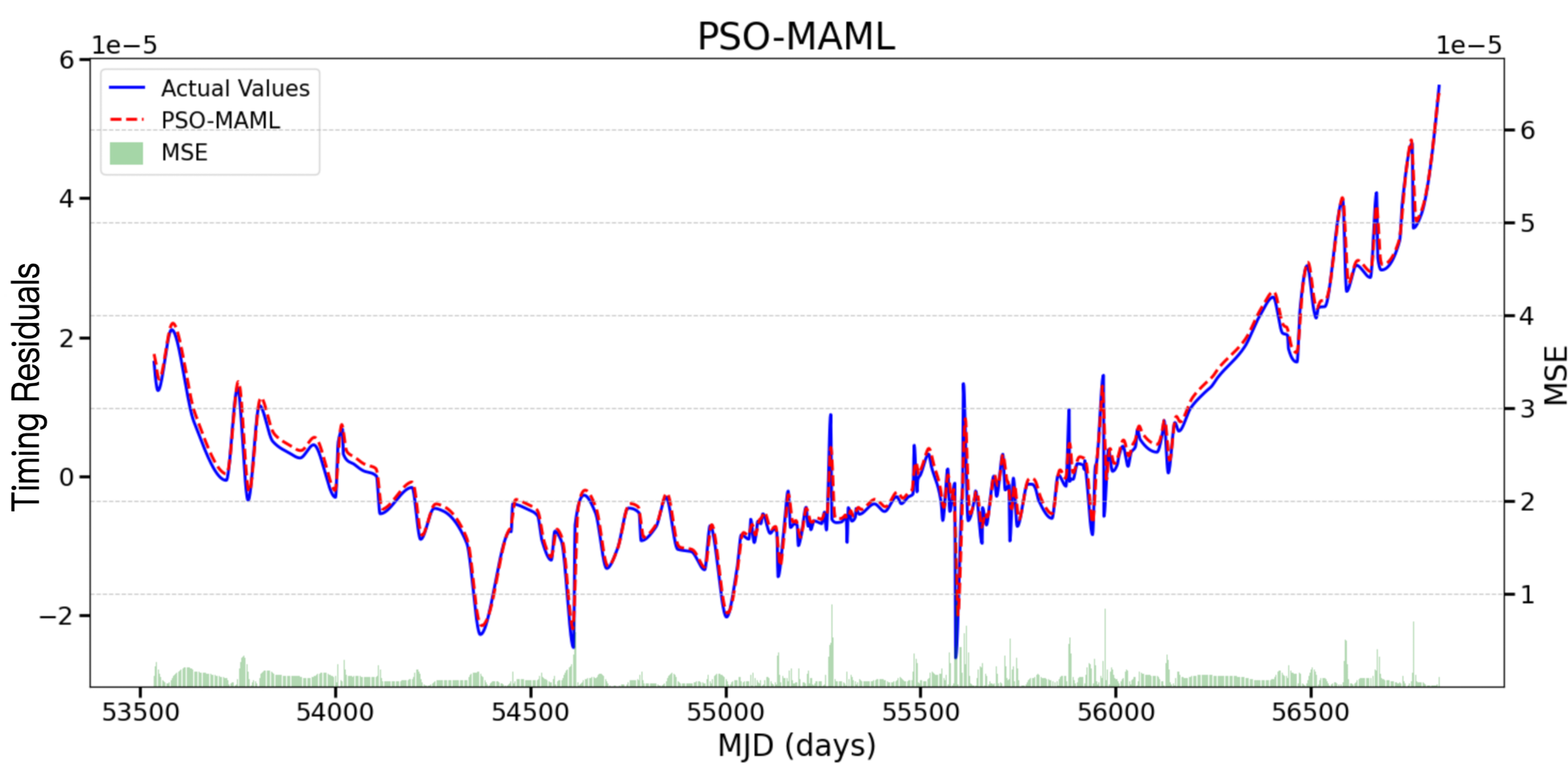} 
    \caption{The predictions of timing residuals at test frequency 541.810 Hz with (black curve) and without (red curve) applying the PSO.}
    \label{fig:PSO}
\end{figure*}

\setlength{\tabcolsep}{23pt}
\begin{table*}[!t]
    \centering
    \caption{The prediction results of timing residuals at three test frequencies, computed from mixed training, transfer learning, or MAML strategies, results are summarized in MAE(\%)\textbar MAPE\textbar $R^2$.}
    \label{tab:MAML_comparative}
    \begin{tabular}{lcccccc}
        \toprule
        Setting & \multicolumn{3}{c}{Frequency (Hz)} \\
        \cmidrule(lr){2-4}
        &  420.189 & 541.810 & 607.678 \\
        \midrule
        Mixed Training & 0.0004\textbar1.0175\textbar0.4241 & 0.0008\textbar1.5254\textbar0.3374 & 0.0006\textbar0.5976\textbar0.6708 \\
        Transfer Learning & 0.0006\textbar3.4075\textbar-0.4155 & 0.0008\textbar2.4771\textbar0.5551 & 0.0008\textbar1.5336\textbar0.3545 \\
        MAML & 0.0001\textbar1.1780\textbar0.9019 & 0.0002\textbar0.8069\textbar0.9789 & 0.0004\textbar0.9602\textbar0.8753 \\
        \bottomrule
    \end{tabular}
\end{table*}

\setlength{\tabcolsep}{23pt}
\begin{table*}[!t]
    \centering
    \caption{The prediction results of timing residuals at three test frequencies with and without applying the PSO algorithm, results are summarized in MAE(\%)\textbar MAPE\textbar $R^2$.}
    \label{tab:PSO}
    \begin{tabular}{lcccccc}
        \toprule
        Setting & \multicolumn{3}{c}{Frequency (Hz)} \\
        \cmidrule(lr){2-4}
        &  420.189 & 541.810 & 607.678 \\
        \midrule
        MAML & 0.0001\textbar1.1780\textbar0.9019 & 0.0002\textbar0.8069\textbar0.9789 & 0.0004\textbar0.9602\textbar0.8753 \\
        PSO-MAML & $0.0000005$\textbar0.2792\textbar0.9820 & 0.0001\textbar0.2021\textbar0.9933 & 0.0002\textbar0.4210\textbar0.9520 \\
        \bottomrule
    \end{tabular}
\end{table*}

The prediction results at test frequency 541.810 Hz, computed from these three settings, are further depicted in \cref{fig:MAML_comparative}, where curves are used to demonstrate the residual predictions and the ground truth, bar chart are for MSE values illustrating the divergence. From \cref{fig:MAML_comparative}, timing residuals predicted by MAML (black curve) effectively captures the dynamics of the ground truth data (blue curve), while the predictions produced by mixed training (green curve) or transfer learning (orange curve) largely deviates from the real values, visualized from the bar chart. These findings demonstrate the strong adaptability of the MAML learning strategy over the learning incompetence with limited data.

\subsection{Evaluating The PSO}
Our solution integrates the PSO algorithm for automatic hyperparameter optimization. To further study the effect of the PSO algorithm, performance on the three test frequency domains are tabulated in \cref{tab:PSO}, showing consistent prediction improvements across all frequency domains with the integration of PSO. For example, the application of PSO achieves 0.2021 MAPE at test frequency 541.810, comparatively lower than the 0.8069 produced without the PSO. Also applying the PSO algorithm leads to increased $R^2$ values across all three test frequencies, such improved capture of systematic data variations validates the PSO's effectiveness.

For further investigation, the prediction results on test frequency 541.810 Hz are illustrated in \cref{fig:PSO}, where the prediction results with the PSO algorithm (red curve) align more closely with the actual residuals (blue curve), fitting both overall trends and local fluctuations more accurately than those produced without the PSO, i.e., MAML solely (black curve).

\begin{figure*}[!htbp]
  \centering
  \begin{subfigure}[b]{1\textwidth}
    \centering
    \includegraphics[width=\textwidth, trim=0 10pt 0 0]{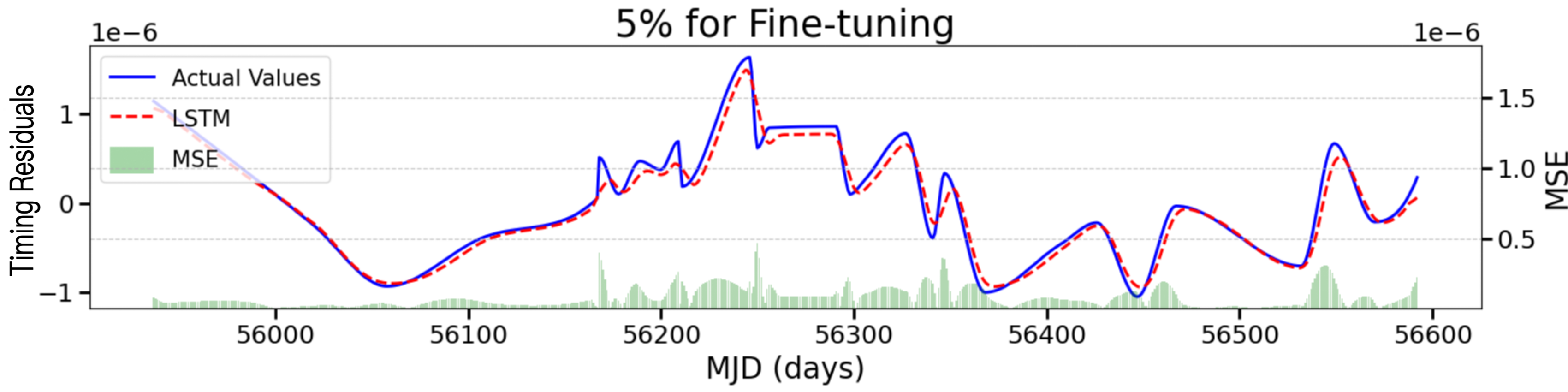}
    \label{fig:resid_vert1}
  \end{subfigure}

  \vspace{-0.5em}
  \begin{subfigure}[b]{1\textwidth}
    \centering
    \includegraphics[width=\textwidth]{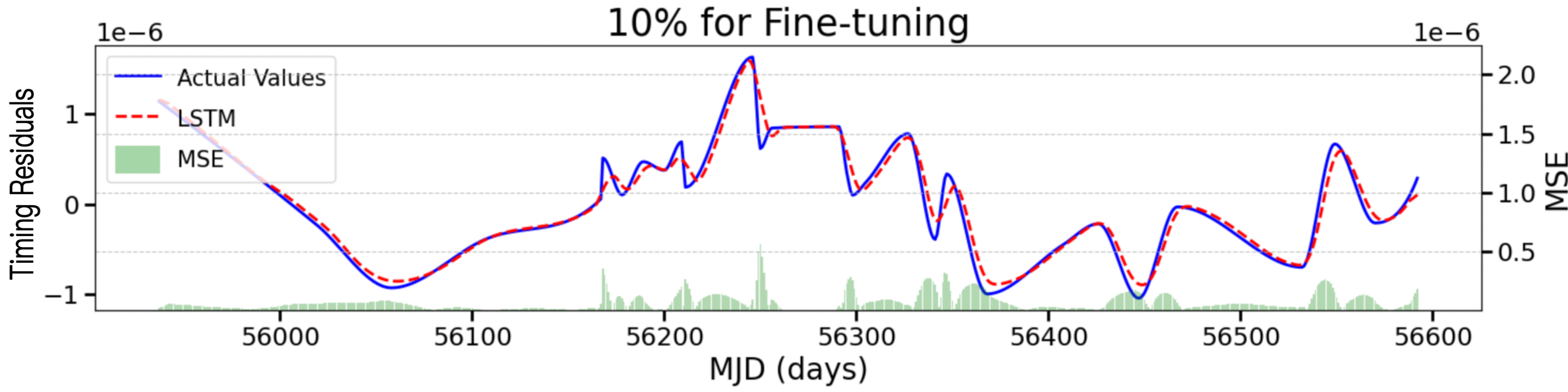}
    \label{fig:resid_vert2}
  \end{subfigure}

  \vspace{-0.5em}
  \begin{subfigure}[b]{1\textwidth}
    \centering
    \includegraphics[width=\textwidth]{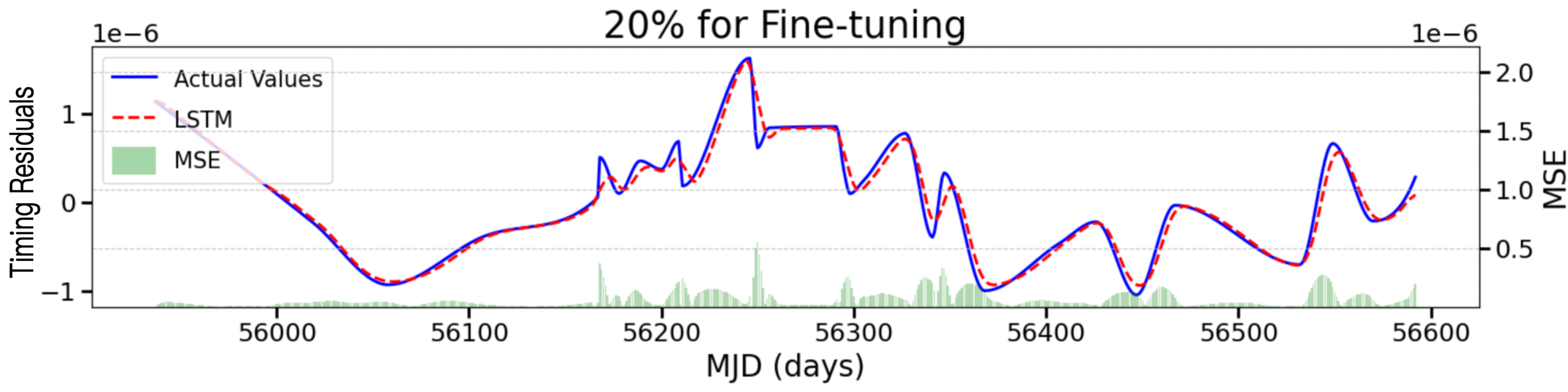}
    \label{fig:resid_vert3}
  \end{subfigure}
  
  \caption{The predictions of timing residuals at test frequency 420.189 Hz by partitioning 5\% (top), 10\% (middle), 20\% (bottom) of data for fine-tuning. Red and blue curves depicts residuals predicted by our solution in full configuration and the real values, respectively.} 
  \label{fig:vertical}
\end{figure*}

\subsection{K-Shot Analysis}
In few-shot learning, the shot number is an important setting that defines the proportion of data used in the MAML's fine-tuning process. In this study, three shot percentiles 5\%, 10\%, and 20\% are selected to fine-tune our meta-trained LSTM, while the last 80\% data is used for evaluation to ensure fair comparisons across the three shot settings. \cref{tab:MAML_shotnumber} tabulates the performance with the three shot numbers at test frequency 420.189 Hz, demonstrating that increasing the fine-tuning data from 5\% to 20\% generally improves prediction results across all three metrics. For instance, the MAE decreases from 0.000007\% to 0.000005\%, while the MAPE declines from 0.3107 to 0.2736. As for the $R^2$, such value rises to $0.9828$ when fine-tuning with 20\% data.

\setlength{\tabcolsep}{10pt}
\begin{table}[!t]
    \centering
    \caption{The prediction results of timing residuals at test frequency 420.189 Hz by varying the shot number.}
    \label{tab:MAML_shotnumber}
    \begin{tabular}{lcccccc}
        \toprule
        Shot-number (\%) & \multicolumn{3}{c}{metrics} \\
        \cmidrule(lr){2-4}
        &  MAE(\%) &  MAPE &  $R^2$ \\
        \midrule
        5 & 0.000007 & 0.3107 & 0.9781 \\
        10 & 0.000005 & 0.2792 & 0.9820 \\
        20 & 0.000005 & 0.2736 & 0.9828 \\
        \bottomrule
    \end{tabular}
\end{table}

In \cref{fig:vertical}, prediction curves at test frequency 420.189 Hz, computed by varying the three shot settings, are depicted to provide additional insights. With 5\% fine-tuning data, the predicted residuals (red curve) deviates from the actual values (blue curve) between MJD 56200 and 56300, suggesting potential underfitting in fine-tuning. As for the 10\% and 20\% settings, the predictions align well with real values and the errors are relatively lower, providing more robust predictions. Despite the best predictions are attained at the 20\% case, we select 10\% for fine-tuning due to the marginal performance difference between these two settings, while the lower proportion of 10\% aligns more closely with our objective of addressing the data scarcity issue. These findings validate our accurate predictions with limited data, also providing valuable insights regarding the proportion of data for fine-tuning the model.

\subsection{Computational Efficiency}
Our solution only contains 13.3M trainable parameters. All the experiments are performed using PyTorch 2.2.1 on a laptop with a 12th Gen Intel Core i9-12900H CPU. Training consumes around 7895.8 MB CPU memory, while single-step residual prediction only costs 16.9 MB CPU memory in 18.0 milliseconds (around 55.5 sample-per-second). The low memory consumption and rapid inference largely improve the practicability, enabling real-time pulsar residual prediction on real-world pulsar observation systems constrained with limited computational resources.

\section{Conclusion}
This work introduces a novel solution for few-shot prediction of pulsar timing residuals, addressing the challenges of data scarcity and model adaptability across domains. Our solution uses an LSTM network optimized with the MAML learning strategy to learn a generalized parameter initialization from low-frequecy domains, enabling rapid and accurate fine-tuning based on limited data from each high-frequency domain. The integration of the PSO algorithm automatically optimizes hyperparameters and therefore makes our solution fully automated with improved prediction results. Key findings from evaluations and experimental analysis on the IPTA dataset are summarized as follows:

\begin{enumerate}[label=\arabic*)]
\item Robust generalization. Our solution demonstrates strong adaptability by enabling neural networks with recurrent layers to operate in data-scarcity scenarios across diverse frequency domains, outperforming traditional transfer learning strategy with reduced errors up to 67.42\% MAPE (e.g., from 2.4771 to 0.8069 on 541.810 Hz).

\item Accurate predictions. Our solution demonstrates accurate prediction results across all tested high-frequency domains, by requiring only 10\% of ground truth data for fine-tuning the model. For example, at the frequency of 541.810 Hz, the solution achieves an MAE of just 0.0001\% and an $R^{2}$ value of 0.9933, demonstrating high precision.

\item Automated solution. The integration of the PSO algorithm enables fully automated optimization of model hyperparameters, allowing the LSTM network to adapt more efficiently to new frequency domains, ensuring rapid and accurate fine-tuning with minimal manual intervention.

\item Exceptional efficiency. Our lightweight structure only consists of 13.3M trainable parameters, requiring only 16.86 MB and 18.0 milliseconds for single-step prediction, making it suitable for real-world applications requiring effective and real-time predictions of pulsar timing residuals in resource-constrained environments.
\end{enumerate}

In conclusion, our work offers an accurate, efficient, and highly adaptable solution for the predictions of pulsar timing residuals in cross-frequency domains, particularly practical in real-world settings. Future research will focus on developing advanced learning strategies to improve predictions in challenging scenarios such as timing noise with weak regularity, therefore further improving our robustness and reliability.

\bibliography{reference.bib}{}

@article{RN03,
  author = {Manchester, R. N. and Hobbs, G. and Bailes, M. and et al.},
  title = {The Parkes Pulsar Timing Array Project},
  journal = {Publications of the Astronomical Society of Australia},
  year = {2013},
  volume = {30},
  pages = {e017}
}

@article{RN04,
  author = {Susobhanan, A. and Kaplan, D. L. and Archibald, A. M. and et al.},
  title = {PINT: Maximum-likelihood estimation of pulsar timing noise parameters},
  journal = {The Astrophysical Journal},
  year = {2024},
  volume = {971},
  number = {2},
  pages = {150}
}

@Article{RN06,
AUTHOR = {Jain, Meetu and Saihjpal, Vibha and Singh, Narinder and Singh, Satya Bir},
TITLE = {An Overview of Variants and Advancements of PSO Algorithm},
JOURNAL = {Applied Sciences},
VOLUME = {12},
YEAR = {2022},
NUMBER = {17},
ARTICLE-NUMBER = {8392},
URL = {https://www.mdpi.com/2076-3417/12/17/8392},
ISSN = {2076-3417}
}

@article{RN07,
  title={Research Progress on X-ray Pulsar Navigation Detectors},
  author={Xiaokun Wang},
  journal={Applied Physics},
  volume={13},
  pages={407},
  year={2023},
  publisher={Chinese Academic Journal, Hans Publishers},
  doi = {https://doi.org/10.12677/APP.2023.1310044},
  
}

@article{RN08,
    author = {Lynch, R S and Swiggum, J K and Kondratiev, V I and Kaplan, D L and Stovall, K and Fonseca, E and Roberts, M S E and DeCesar, M E and Levin, L and Rosen, R and others},
    title = {The Green Bank North Celestial Cap Pulsar Survey. III. 45 New Pulsar Timing Solutions},
    journal = {The Astrophysical Journal},
    volume = {859},
    number = {2},
    pages = {93},
    year = {2018},
    month = {05},
    doi = {10.3847/1538-4357/aabf8a},
    url = {https://ui.adsabs.harvard.edu/link_gateway/2018ApJ...859...93L/doi:10.3847/1538-4357/aabf8a}
}

@article{RN09,
  author = {Perera, B B P and DeCesar, M E and Demorest, P B and Kerr, M and Lentati, L and Nice, D J and Os{\l}owski, S and Ransom, S M and Keith, M J and Arzoumanian, Z and others},
    title = {The International Pulsar Timing Array: second data release},
    journal = {Monthly Notices of the Royal Astronomical Society},
    volume = {490},
    number = {4},
    pages = {4666-4687},
    year = {2019},
    month = {12},
    issn = {0035-8711},
    doi = {10.1093/mnras/stz2857},
    url = {https://doi.org/10.1093/mnras/stz2857}
}

@article{RN10,
  title={Enhancing Pulsar Candidate Identification with Self-tuning Pseudolabeling Semisupervised Learning},
  author={Liu, Yi and Jin, Jing and Zhao, Hongyang and Wang, Zhenyi},
  journal={The Astrophysical Journal},
  volume={967},
  number={2},
  pages={155},
  year={2024},
  publisher={IOP Publishing}
}

@article{RN14,
  title = {Long short-term memory},
  author = {Hochreiter, S. and Schmidhuber, J.},
  journal = {Neural Computation},
  volume = {9},
  number = {8},
  pages = {1735-1780},
  year = {1997},
}

@article{RN15,
  title={Long short-term memory},
  author={Graves, Alex and Graves, Alex},
  journal={Supervised sequence labelling with recurrent neural networks},
  pages={37--45},
  year={2012},
  publisher={Springer}
}

@article{RN16,
  title={Particle swarm optimization algorithm: an overview},
  author={Wang, Dongshu and Tan, Dapei and Liu, Lei},
  journal={Soft computing},
  volume={22},
  number={2},
  pages={387--408},
  year={2018},
  publisher={Springer}
}

@article{RN17,
  author  = {Antoniadis, J. and Babak, S. and Nielsen, A. S. B. and {et al.}},
  title   = {The Second Data Release from the European Pulsar Timing Array: The Dataset and Timing Analysis},
  journal = {Astronomy \& Astrophysics},
  year    = {2023},
  volume  = {678},
  pages   = {A48},
}

@article{RN18,
  author  = {Arzoumanian, Z. and Brazier, A. and Burke-Spolaor, S. and {et al.}},
  title   = {The NANOGrav Nine-year Data Set: Observations, Arrival Time Measurements, and Analysis of 37 Millisecond Pulsars},
  journal = {The Astrophysical Journal},
  year    = {2015},
  volume  = {813},
  number  = {1},
  pages   = {65},
}

@article{RN19,
  author  = {Zhu, Xingjiang},
  title   = {The Parkes Pulsar Timing Array},
  journal = {Physics (China)},
  year    = {2024},
  volume  = {53},
  number  = {8},
  pages   = {525--531},
  language= {Chinese},
}

@inproceedings{RN20,
  title={PulChron: A pulsar time scale demonstration for PNT systems},
  author={P{\'\i}riz, Ricardo and Garbin, Esteban and Rold{\'a}n, Pedro and Keith, Michael and Shaw, Benjamin and Shemar, Setnam and Burrows, Kathryn and Davis, John and Binda, Stefano},
  booktitle={Proceedings of the 50th annual precise time and time interval systems and applications meeting},
  pages={191--205},
  year={2019}
}

@article{RN21,
  title={An improved wiener filtration method for constructing the ensemble pulsar timescale},
  author={Zhang, Zhehao and Tong, Minglei and Yang, Tinggao},
  journal={The Astrophysical Journal},
  volume={962},
  number={1},
  pages={2},
  year={2024},
  publisher={IOP Publishing}
}

@article{RN22,
  title={Timing of young radio pulsars--I. Timing noise, periodic modulation, and proper motion},
  author={Parthasarathy, A and Shannon, RM and Johnston, S and Lentati, L and Bailes, Matthew and Dai, S and Kerr, M and Manchester, RN and Os{\l}owski, S and Sobey, C and others},
  journal={Monthly Notices of the Royal Astronomical Society},
  volume={489},
  number={3},
  pages={3810--3826},
  year={2019},
  publisher={Oxford University Press}
}

@article{RN23,
  title={Assessing the role of spin noise in the precision timing of millisecond pulsars},
  author={Shannon, Ryan M and Cordes, James M},
  journal={The Astrophysical Journal},
  volume={725},
  number={2},
  pages={1607},
  year={2010},
  publisher={IOP Publishing}
}

@article{RN24,
  title={Arrival-time analysis for a millisecond pulsar},
  author={Blandford, Roger and Narayan, Ramesh and Romani, Roger W},
  journal={Journal of Astrophysics and Astronomy},
  volume={5},
  pages={369--388},
  year={1984},
  publisher={Springer}
}

@article{RN25,
  title={Precision timing of PSR J0437--4715: an accurate pulsar distance, a high pulsar mass, and a limit on the variation of Newton’s gravitational constant},
  author={Verbiest, Joris PW and Bailes, Matthew and Van Straten, W and Hobbs, George B and Edwards, Russell T and Manchester, Richard N and Bhat, NDR and Sarkissian, John M and Jacoby, Bryan A and Kulkarni, Shri R},
  journal={The Astrophysical Journal},
  volume={679},
  number={1},
  pages={675},
  year={2008},
  publisher={IOP Publishing}
}

@article{RN27,
  title={The international pulsar timing array: first data release},
  author={Verbiest, JPW and Lentati, L and Hobbs, G and van Haasteren, Rutger and Demorest, Paul B and Janssen, GH and Wang, J-B and Desvignes, G and Caballero, RN and Keith, MJ and others},
  journal={Monthly Notices of the Royal Astronomical Society},
  volume={458},
  number={2},
  pages={1267--1288},
  year={2016},
  publisher={The Royal Astronomical Society}
}

@ARTICLE{RN28,
       author = {{Kaspi}, V.~M. and {Taylor}, J.~H. and {Ryba}, M.~F.},
        title = "{High-Precision Timing of Millisecond Pulsars. III. Long-Term Monitoring of PSRs B1855+09 and B1937+21}",
      journal = {\apj},
         year = 1994,
        month = jun,
       volume = {428},
        pages = {713},
          doi = {10.1086/174280},
       adsurl = {https://ui.adsabs.harvard.edu/abs/1994ApJ...428..713K}
}

@article{RN29,
  title={Testing theories of gravitation using 21-year timing of pulsar binary J1713+ 0747},
  author={Zhu, WW and Stairs, IH and Demorest, PB and Nice, David J and Ellis, JA and Ransom, SM and Arzoumanian, Z and Crowter, K and Dolch, T and Ferdman, RD and others},
  journal={The Astrophysical Journal},
  volume={809},
  number={1},
  pages={41},
  year={2015},
  publisher={IOP Publishing}
}

@article{RN30,
  title={From spin noise to systematics: stochastic processes in the first International Pulsar Timing Array data release},
  author={Lentati, Lindley and Shannon, Ryan M and Coles, William A and Verbiest, Joris PW and van Haasteren, Rutger and Ellis, JA and Caballero, RN and Manchester, Richard Norman and Arzoumanian, Z and Babak, Stanislav and others},
  journal={Monthly Notices of the Royal Astronomical Society},
  volume={458},
  number={2},
  pages={2161--2187},
  year={2016},
  publisher={The Royal Astronomical Society}
}

@article{RN31,
  title={Assessing the role of spin noise in the precision timing of millisecond pulsars},
  author={Shannon, Ryan M and Cordes, James M},
  journal={The Astrophysical Journal},
  volume={725},
  number={2},
  pages={1607},
  year={2010},
  publisher={IOP Publishing}
}

@article{RN32,
  title={Limitations in timing precision due to single-pulse shape variability in millisecond pulsars},
  author={Shannon, Ryan M and Os{\l}owski, S and Dai, S and Bailes, Matthew and Hobbs, G and Manchester, Richard Norman and van Straten, Willem and Raithel, Carolyn A and Ravi, Vikram and Toomey, L and others},
  journal={Monthly Notices of the Royal Astronomical Society},
  volume={443},
  number={2},
  pages={1463--1481},
  year={2014},
  publisher={Oxford University Press}
}

@inproceedings{RN33,
  title={Attention-based bidirectional long short-term memory networks for relation classification},
  author={Zhou, Peng and Shi, Wei and Tian, Jun and Qi, Zhenyu and Li, Bingchen and Hao, Hongwei and Xu, Bo},
  booktitle={Proceedings of the 54th annual meeting of the association for computational linguistics (volume 2: Short papers)},
  pages={207--212},
  year={2016}
}

@inproceedings{RN34,
  title={Bidirectional long short-term memory networks for relation classification},
  author={Zhang, Shu and Zheng, Dequan and Hu, Xinchen and Yang, Ming},
  booktitle={Proceedings of the 29th Pacific Asia conference on language, information and computation},
  pages={73--78},
  year={2015}
}

@article{RN35,
  title={Inceptiontime: Finding alexnet for time series classification},
  author={Ismail Fawaz, Hassan and Lucas, Benjamin and Forestier, Germain and Pelletier, Charlotte and Schmidt, Daniel F and Weber, Jonathan and Webb, Geoffrey I and Idoumghar, Lhassane and Muller, Pierre-Alain and Petitjean, Fran{\c{c}}ois},
  journal={Data Mining and Knowledge Discovery},
  volume={34},
  number={6},
  pages={1936--1962},
  year={2020},
  publisher={Springer}
}

@article{RN36,
  title={Generative adversarial nets},
  author={Goodfellow, Ian J and Pouget-Abadie, Jean and Mirza, Mehdi and Xu, Bing and Warde-Farley, David and Ozair, Sherjil and Courville, Aaron and Bengio, Yoshua},
  journal={Advances in neural information processing systems},
  volume={27},
  year={2014}
}

@article{RN37,
  title={Denoising diffusion probabilistic models},
  author={Ho, Jonathan and Jain, Ajay and Abbeel, Pieter},
  journal={Advances in neural information processing systems},
  volume={33},
  pages={6840--6851},
  year={2020}
}

@inproceedings{RN38,
  title={Model-agnostic meta-learning for fast adaptation of deep networks},
  author={Finn, Chelsea and Abbeel, Pieter and Levine, Sergey},
  booktitle={International conference on machine learning},
  pages={1126--1135},
  year={2017},
  organization={PMLR}
}

@inproceedings{RN39,
  title={Particle swarm optimization for hyper-parameter selection in deep neural networks},
  author={Lorenzo, Pablo Ribalta and Nalepa, Jakub and Kawulok, Michal and Ramos, Luciano Sanchez and Pastor, Jos{\'e} Ranilla},
  booktitle={Proceedings of the genetic and evolutionary computation conference},
  pages={481--488},
  year={2017}
}

@article{RN40,
  title={Assessing the role of spin noise in the precision timing of millisecond pulsars},
  author={Shannon, Ryan M and Cordes, James M},
  journal={The Astrophysical Journal},
  volume={725},
  number={2},
  pages={1607},
  year={2010},
  publisher={IOP Publishing}
}

@article{RN41,
  title={Pulsar timing analysis in the presence of correlated noise},
  author={Coles, W and Hobbs, G and Champion, DJ and Manchester, RN and Verbiest, JPW},
  journal={Monthly Notices of the Royal Astronomical Society},
  volume={418},
  number={1},
  pages={561--570},
  year={2011},
  publisher={Blackwell Publishing Ltd Oxford, UK}
}

@article{RN42,
  title={Hyper-efficient model-independent Bayesian method for the analysis of pulsar timing data},
  author={Lentati, Lindley and Alexander, Paul and Hobson, Michael P and Taylor, Stephen and Gair, Jonathon and Balan, Sreekumar T and van Haasteren, Rutger},
  journal={Physical Review D—Particles, Fields, Gravitation, and Cosmology},
  volume={87},
  number={10},
  pages={104021},
  year={2013},
  publisher={APS}
}

@article{RN43,
  title={Understanding and analysing time-correlated stochastic signals in pulsar timing},
  author={van Haasteren, Rutger and Levin, Yuri},
  journal={Monthly Notices of the Royal Astronomical Society},
  volume={428},
  number={2},
  pages={1147--1159},
  year={2013},
  publisher={Oxford University Press}
}

@article{RN44,
  title={Analysing radio pulsar timing noise with a Kalman filter: a demonstration involving PSR J1359- 6038},
  author={O’Neill, Nicholas J and Meyers, Patrick M and Melatos, Andrew},
  journal={Monthly Notices of the Royal Astronomical Society},
  volume={530},
  number={4},
  pages={4648--4664},
  year={2024},
  publisher={Oxford University Press}
}

@article{RN46,
  title={The International Pulsar Timing Array second data release: Search for an isotropic gravitational wave background},
  author={Antoniadis, John and Arzoumanian, Z and Babak, S and Bailes, M and Bak Nielsen, AS and Baker, P\_T and Bassa, C\_G and B{\'e}csy, B and Berthereau, A and Bonetti, M and others},
  journal={Monthly Notices of the Royal Astronomical Society},
  volume={510},
  number={4},
  pages={4873--4887},
  year={2022},
  publisher={Oxford University Press}
}

@article{dataset,
  title={The international pulsar timing array: first data release},
  author={Verbiest, JPW and Lentati, L and Hobbs, G and van Haasteren, Rutger and Demorest, Paul B and Janssen, GH and Wang, J-B and Desvignes, G and Caballero, RN and Keith, MJ and others},
  journal={Monthly Notices of the Royal Astronomical Society},
  volume={458},
  number={2},
  pages={1267--1288},
  year={2016},
  publisher={The Royal Astronomical Society}
}

@article{RN47,
  author       = {Ruo-Yu Sun},
  title        = {Optimization for Deep Learning: An Overview},
  journal      = {Journal of the Operations Research Society of China},
  volume       = {8},
  pages        = {249--294},
  year         = {2020},
  doi          = {10.1007/s40305-020-00309-6}
}

@article{RN48,
  title={Timing properties of PSR 1951+ 32 in the CTB 80 supernova remnant},
  author={Foster, RS and Backer, DC and Wolszczan, A},
  journal={Astrophysical Journal, Part 1 (ISSN 0004-637X), vol. 356, June 10, 1990, p. 243-249.},
  volume={356},
  pages={243--249},
  year={1990}
}

@article{RN49,
  title={Opportunities for detecting ultralong gravitational waves},
  author={Sazhin, Mikhail V},
  journal={Sov. Astron},
  volume={22},
  pages={36--38},
  year={1978}
}

@article{RN50,
  title={Black holes and gravitational waves. II-Trajectories plunging into a nonrotating hole},
  author={Detweiler, Steven L and Szedenits Jr, Eugene},
  journal={Astrophysical Journal, Part 1, vol. 231, July 1, 1979, p. 211-218.},
  volume={231},
  pages={211--218},
  year={1979}
}

@article{RN52,
  title={The NANOGrav 15 yr data set: observations and timing of 68 millisecond pulsars},
  author={Agazie, Gabriella and Alam, Md Faisal and Anumarlapudi, Akash and Archibald, Anne M and Arzoumanian, Zaven and Baker, Paul T and Blecha, Laura and Bonidie, Victoria and Brazier, Adam and Brook, Paul R and others},
  journal={The Astrophysical Journal Letters},
  volume={951},
  number={1},
  pages={L9},
  year={2023},
  publisher={IOP Publishing}
}

@article{R201,
   title={Understanding and analysing time-correlated stochastic signals in pulsar timing},
   volume={428},
   ISSN={0035-8711},
   url={http://dx.doi.org/10.1093/mnras/sts097},
   DOI={10.1093/mnras/sts097},
   number={2},
   journal={Monthly Notices of the Royal Astronomical Society},
   publisher={Oxford University Press (OUP)},
   author={van Haasteren, Rutger and Levin, Yuri},
   year={2012},
   month=oct, pages={1147–1159} }

@article{R202,
  title={New advances in the Gaussian-process approach to pulsar-timing data analysis},
  author={van Haasteren, Rutger and Vallisneri, Michele},
  journal={Physical Review D},
  volume={90},
  number={10},
  pages={104012},
  year={2014},
  publisher={APS}
}
\bibliographystyle{aasjournal}

\end{document}